\documentclass[runningheads]{llncs}
\usepackage{graphicx}
\usepackage{biblatex}
\begin{document}
\title{Share with Me: A Study on a Social Robot Collecting Mental Health Data\thanks{This study was funded in part by National Science Foundation: National Robotics Initiative, SES: Award Abstract 1734100 - Design and Development of a Social Robot for Gathering Ecological Momentary Stress Data from Teens.}}
%
%
\author{Raida Karim\inst{1}\orcidID{0000-0002-2854-3985} \and
Edgar Lopez\inst{1}\orcidID{0000-0001-6835-855X} \and
Katelynn Oleson\inst{1} \and
Tony Li\inst{1}\orcidID{0000-0001-7552-6689} \and
Elin A. Björling\inst{1}\orcidID{0000-0002-0385-2562} \and
Maya Cakmak\inst{1}\orcidID{0000-0001-8457-6610}}
\authorrunning{R. Karim et al.}
%
\institute{University of Washington WA 98195, USA \\
\email{\{rk1997, mcakmak\}@cs.washington.edu, \\
\{lopeze7, kjoleson, tonywli, bjorling\}@uw.edu}}
\maketitle              
\begin{abstract}
Social robots have been used to assist with mental well-being in various ways such as to help children with autism improve on their social skills and executive functioning such as joint attention and bodily awareness. They are also used to help older adults by reducing feelings of isolation and loneliness, as well as supporting mental well-being of teens and children. However, existing work in this sphere has only shown support for mental health through social robots by responding interactively to human activity to help them learn relevant skills. We hypothesize that humans can also get help from social robots in mental well-being by releasing or sharing their mental health data with the social robots. In this paper, we present a human-robot interaction (HRI) study to evaluate this hypothesis. During the five-day study, a total of fifty-five (n=55) participants shared their in-the-moment mood and stress levels with a social robot. We saw a majority of positive results indicating it is worth conducting future work in this direction, and the potential of social robots to largely support mental well-being. 

\keywords{Social Robot  \and Mental Health \and Data Sharing.}
\end{abstract}
\section{Introduction}
Apart from genetic or birth related causes, any kind of mental health issues in our daily lives are usually caused by some kind of trouble such as stressful events, grief from accidents or deaths, and trauma from tragic or fearful incidents~\cite{schomerus2006public}. Research shows that sharing these troubling experiences helps gain insights from others which helps develop coping skills, and makes one feel less alone in pain which helps to tackle trouble more effectively~\cite{facebook}\cite{moodwall}. Therefore, sharing about trouble can potentially help treat or lessen mental health issues.
However, people are not always willing to share their mental health with others or publicly, even with their family members, let alone outside community members due to social stigma, personal beliefs or limitations~\cite{SARTORIUS2007810}. Therefore, to tackle mental health issues, the main source of support is usually individualized therapy, which is expensive and thus inaccessible to many~\cite{brown2017discussing}\cite{pella2020pediatric}. Digital therapy is a relatively more accessible option compared to in-person therapy because of lower cost and availability in any location~\cite{aboujaoude2020digital}\cite{taylor2020digital}. However, digital therapy has drawbacks, such as lack of face-to-face interaction, user disengagement, and software incompatibility~\cite{digthe}. So, we hypothesize that sharing mental health with an endearing social robot that overcomes these challenges of therapy options can help support mental health.

\section{Related Work}
When people share their personally relevant emotions in social media like Facebook, they experience satisfaction causing positive mental state~\cite{facebook}. People tend to be discreet about sharing their emotional data, as we see Facebook users share more intense and negative emotions in private messages~\cite{facebook}. Prior work that looked into sharing one's emotions through technologies like a virtual mood wall reported a positive relation between online emotional sharing and comfort in negative emotions~\cite{moodwall}. Social robots are often perceived as friendly or pet-like companions by humans for their endearing appearances (e.g., outfit, facial expressions) and capabilities that can include haptics, sounds, and movements~\cite{robot}. Such perceptions or relationships boost user engagement with social robots~\cite{bishop}. Thus, social robots can better foster emotional support with mental health data compared to other technologies (e.g., smartphone apps, websites rendered by touch screen devices)~\cite{boucenna2014interactive}. In a study conducted by Björling et al.~\cite{elin}, teens reported social robots as a plausible source of emotional support. Additionally, social robots have been studied in the context of assisting mental healthcare by sharing its emotions with people~\cite{sharestress}, having behavioral models fulfilling user needs~\cite{robotics8030054}, and providing companionship~\cite{inbook}.
\newline
To the best of our knowledge, the existing literature in this domain does indicate that social robots have not yet been used as a means for collecting mental health data from humans, which makes this research direction a novel one in HRI.

\section{Sharing Mental Health Data with a Social Robot}
\subsection{Study Setup}
We conduct a five-day study with some people showing to share their mental health data with a social robot, and based on their reactions, we analyzed the potential of such interactions in improving mental well-being.
The goal of this study is to evaluate our hypothesis as stated before. The research questions are:\\
(A) How will participants feel about sharing their mental health data with a social robot? \\
(B) What information can we get about a community's mental health from its participants' shared mental health data with a social robot?
\\
We plan to investigate these research questions with findings from the study results as discussed later.
\begin{figure}
\includegraphics[width=\textwidth]{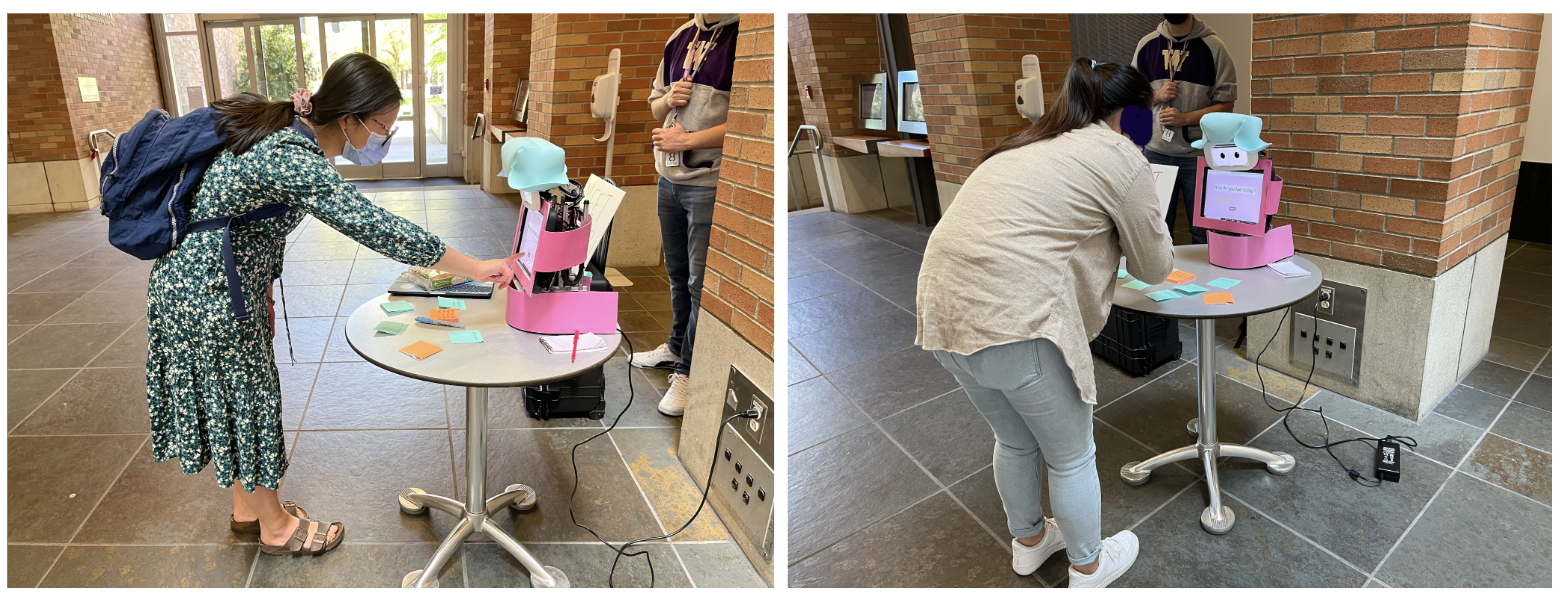}
\caption{Participants sharing mental health data with a social robot during the study.} \label{fig:study}
\end{figure}
\subsection{Sample}
We conducted the study with a total of fifty-five (n=55) participants in an American university campus, specifically a top Computer Science \& Engineering (CSE) school. The location chosen was the main entrance atrium of the CSE department easily accessible to students and staff to interact with
the robot. Participants interacted with a social robot shown in Fig.~\ref{fig:study} to share their mental health data in this five days’ study.
\subsection{Procedure}
We asked the participants to share their mood and stress levels as a means to share their mental health data with a social robot. A previous study conducted by our lead author~\cite{lbr} showed that rather than using English words like “Good”, “Bad”, and “Fine” to express different mood and stress levels, using an emoji Likert scale can enhance coherence and accessibility by reducing the need to read text and increase dependency on globally understandable graphics and symbols, especially in diverse communities like in the United States. Thus we used an emoji likert scale with 3 buttons expressing 3 different levels of mood or stress data.
\newline
Before performing the study in the CSE department, we conducted a pilot study\footnote{A video of the pilot study of the mental health data sharing interaction by one participant with a social robot can be found here: \url{https://youtu.be/J6srKDg6OE0}} with a female participant in order to ensure that the planned experimental interaction can help us find out more about the participants' shared mental health data as expected. The brief (about 20 seconds) interaction had 3 stages: greeting, mood and stress data sharing, and exit. In the data sharing or second stage, participants were asked to input their \textit{mood} and \textit{stress} data by clicking on the appropriate emoji as answers to these two questions, respectively: \textit{“How do you feel today?”} and \textit{“Are you stressed today?”} Their shared data responses were then stored in a secured Firebase database \footnote{Firebase: \url{https://firebase.google.com/}}. We decorated the robot with a colorful hat and body frame to make it look endearing (see Fig.~\ref{fig:study}). Apart from showing the questions in its belly screen, there were sounds incorporated in the robot that asked the questions aloud in a female voice in English language. The robot also had head and neck movements in different stages of the interaction. The first three authors of this paper invited passerby on different days verbally to participate in the study. After each passerby completed the interaction, they were interviewed to respond either verbally or by writing in post-its to a prompt as shown in Fig.~\ref{fig:prompts}. 

\subsection{Results}
\begin{figure}
\centering
\includegraphics[width=\textwidth]{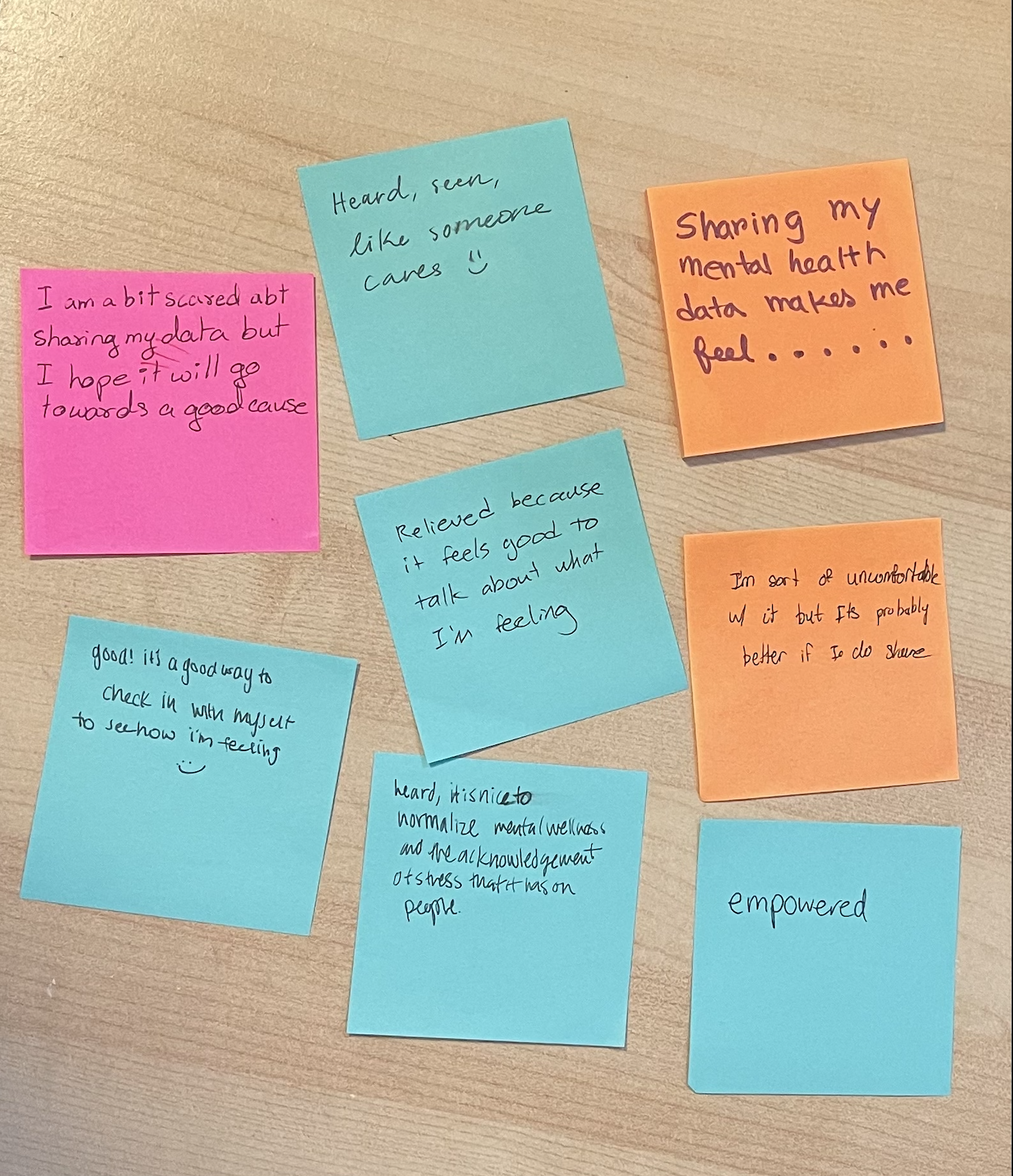}
\caption{Written Prompt Used in the Study \& Responses in Post-its.} \label{fig:prompts}
\end{figure}

The results of the study include participants’ comments responding to the prompt of \textit{“Sharing my mental health data makes me feel….”} (see Fig.~\ref{fig:prompts}) and the shared mood and stress level data through emoji counts. We observed in these data that a positive mood was prevalent on Monday, Tuesday, Thursday, and Friday. A neutral mood was prevalent on Wednesday and Friday. A negative mood was not prevalent in this community on any days in that week. In terms of stress data, high stress was prevalent on Wednesday, medium stress was prevalent on Monday, Tuesday, Thursday, and Friday, and low stress was prevalent on Thursday. Of course, these observations are true for that specific week, and might not necessarily hold true for other weeks. We saw the lowest positive comments on Friday, which may mean this community is exhausted after a long busy week, and thus did not think much or see many positive aspects of this new type of interaction.
\subsection{Analysis}
The lead author of this paper conducted a thematic analysis of the study data containing prompt responses of the study participants as shown in Fig.~\ref{fig:prompts}. A specific theme was assigned based on the nature of prompt response. For instance, comments like \textit{“Sharing my mental health data makes me feel more aware of my stress.”} has been assigned to the theme of “self-awareness”. And comments like \textit{“I am a bit scared about sharing my data but hope it will go towards a good cause.”} has been assigned to the theme of “privacy” as it shows concerns about how participants’ data will be used and possible risks to their privacy.
\newline
 The themes are mostly positive supporting our hypothesis that sharing mental health data with a social robot can support mental healthcare. The only negative themes involved privacy concerns about data sharing and feedback for improvement on robot design. We can mitigate these negative themes by incorporating participants’ feedback on the robot’s design and ensuring participants about the privacy of their shared data in our secured data storage. One observation we made was that this community is part of an academically competitive department, so these results seem to align with this competitive mentality. For instance, from overall community counts, it seems this community most commonly has a good or positive mood and medium stress. Good mood could be a result of the satisfaction and achievement that comes with being part of a prestigious academic department, and being able to pursue the profession they are very passionate about in a competitive place. The medium stress might result from the busy nature of academic or engineering work-- deadlines of homework, projects, papers, presentations, demos, etc., that all are constantly working on.

We discuss specific themes below:
\newline
\textbf{Theme 1: Emotional Sharing}
\newline
Emotions revealed by users' post-interaction comments ranged between positive (e.g., \textit{``happy"}, \textit{``good, could be myself"}), unchanged (e.g., \textit{``Neutral. No change in how I feel"}, \textit{``didn't feel anything"}), and validated (e.g., \textit{``Accepted"}, \textit{``welcomed, nice to check-in with someone"}, \textit{``Heard, seen, like someone cares :)"}).
\newline
\underline{Implications:}
This theme reinforces that sharing humans' mental health data with a social robot can indeed have various effects in their emotions. These findings can inform future designs of data sharing interaction with social robots targeted to impact the users' mental health for the better.
\newline
\textbf{Theme 2: Robot Interaction}
\newline
Some users expressed how they felt interacting with a robot specifically while sharing their mental health data. One type of comments expressed the robot's figure positively impacting mental health: \textit{``I feel relaxed by the face", ``It made my day to share with the robot"}.
\newline
Another type of comments expressed doubts in the robot's capability: \textit{``As it's a robot, I know that it will lack human level empathy. The set of responses would be fixed. I knew it won't be capable of doing much like releasing my stress.", ``confused, not sure what happened or what's happening here, does it have voice and movements"}.
\newline
\underline{Implications:} These findings indicate the need to design or use social robots that can transparently and positively support mental health data sharing interaction. In general, the term “creepy technology” is increasingly appearing in literature along with concerns for ethical boundaries, cybersecurity, privacy, and mistaken identity~\cite{McWhorter2020CreepyTA}. Creepiness has been noted to have three core factors– implied malice, undesirability, and unpredictability– in the first work that devoted detailed attention to creepiness in the context of interactive technologies~\cite{Wo_niak_2021}. Therefore, when designing these interactions, ensuring transparency, desirability, predictability, and privacy can help avoid the robot being perceived as creepy.
\newline
\textbf{Theme 3: Inward-Looking}
\newline
This theme is comprised of comments that show users contemplating themselves during the interaction. Some keywords users conveyed: \textit{``reflective", ``Transparent", ``introspective"}.
\newline
\underline{Implications:} These findings suggest this interaction's ability to compel humans for self-reflection, which can inform future design of such interactions to allow space for reflection. 
\newline
\textbf{Theme 4: Impact on Stress Level}
\newline
This theme reveals how this interaction impacts users' stress levels. Some comments demonstrated alleviation in stress: \textit{``Less Stressed", ``like some weight has been lifted off of me"}, whereas one comment shows no change in stress level: \textit{``same level of stress"}.
\newline
\underline{Implications:} These findings show potential for this type of interaction as a means of alleviating stress. Future research needs to be conducted to solidify this possibility. Diverse communities, targeted interaction experiments, and a larger number of participants can be used to conduct further research in different settings.  
\newline
\textbf{Theme 5: Community Benefit}
\newline
This theme suggests how this interaction could be beneficial to the community. One comment towards this: \textit{``helpful to others"}. 
\newline
\underline{Implications:} These findings point out the need to conduct further research to evaluate how this interaction can be beneficial for people or communities. Depending on in what ways these can be beneficial, the design of such interactions or used social robots could be modified for maximum benefit in specific contexts.  
\newline
\textbf{Theme 6: Feedback}
\newline
This theme contains comments by users as suggestions for improvement. Some comments on interactions: \textit{``weird...as I'm not sure if my mental health can be quantified by data alone", ``unusual"}, and one comment on robot design: \textit{``It didn't feel like a robot. Maybe one integrated system might make it feel like a robot. A frame around the belly might help hide the belly ipad, and make it look like an integrated system. But, you can think more about it like arms coming out, or to make it more humanoid; maybe remove distance between face and belly. If the question appears in the face and I answer in the belly, it might be more like an integrated or full system interaction."}
\newline
\underline{Implications:} These findings indicate some concrete areas for improvement in this interaction and robot design. In general, mental health is a bit abstract and it can be hard to design convincing short (e.g., 20-seconds) interactions to collect accurate mental health data. Incorporating these feedback and conducting further research might lead to new developments in this regard.
\newline
\textbf{Theme 7: Self-Awareness}
\newline
This theme shows user comments with regards to increased awareness in various ways, such as: \textit{``more aware of my stress", ``more conscious about my overall health, since mental health is even more important than physical health in my opinion :)", ``I feel more self-aware"}.
\newline
\underline{Implications:} These findings are promising in the context of using this type of interactions in different communities as a means of raising awareness of mental health care or fighting associated stigma.
\newline
\textbf{Theme 8: Privacy}
\newline
This theme indicates users' feelings when sharing their ``private" mental health data in this interaction. Some comments didn't express any concerns: \textit{``I typically don't like sharing my data, but I liked this one. I liked that there were options given that I could choose from, so I didn't have to think a lot about my mood, etc.", ``I am not concerned at all about sharing my data this way, as it seems low granularity. Not invasive."} Some other comments showed users' concerns: \textit{``I am a bit scared about sharing my data but hope it will go towards a good cause.", ``I'm sort of uncomfortable w/ it but it's probably better if I do share."}
\newline
\underline{Implications:} These findings demonstrate the need for further work in making users feel comfortable and gaining their trust with their private data shared with a robot. Establishing a good relationship with users may increase effectiveness of such interactions.
\section{Limitations}
As shown in Fig.~\ref{fig:study}, the robot used in this study had a belly and a face covered with external outfits which made it look a bit like a humanoid robot~\cite{humanoid} perhaps making participants feel they are sharing their mental health data with a human-like companion or friend. Prior work showed humans prefer interacting with human-looking robots~\cite{humano}. So, the results of this study might be biased towards this type of robot compared to that of a non-humanoid robot design like a Knightscope robot~\cite{wiggers_2017}. Therefore, based on robot designs this study results might vary, and thus we cannot claim this study's inferences are true for all social robots.
Another potential limitation is data bias. As we collected data through the robot placed in an open space of a community, this might have caused data bias. For instance, if the user is aware that the study coordinators or other community members can observe them when they are entering their data, they might not be truthful (especially if it's negative or sensitive mental health data, they might feel hesitation or shame to share that in public). This could be the reason why we didn’t get very high counts of negative data (high stress, negative mood, or negative comments about mental health). One way to avoid this bias might be to experiment with data collection in a closed space. For example, placing the robot in a closed room, where only the user can enter to share their data, or placing a cover around the belly of the robot that will not allow the user's interaction (or at least what data they are entering) to be seen. 
\newline
Lastly, the medium of data collection can pose limitations. Instead of using an emoji Likert scale, if we had used a 10-pt slider scale on a 0-9 range, we might have more nuanced data of mood and stress levels indicated by specific values (e.g., 7) as well as provided participants with more options to accurately express their mental health data.
\section{Conclusion \& Future Directions}
To conclude, our study findings revealed mostly positive feelings about this type of human-robot mental health data sharing interaction among participants, as well as different arousal this interaction can have on people such as self-awareness, emotional sharing, etc. as discussed in the ``Analysis" section above. The majority of positive results indicate such interactions with social robots need to be studied further in different settings to evaluate their potential in mental healthcare holistically. Future studies can be conducted with some changed variables in current study settings such as a different location (e.g., hospital, high schools, public library, online), population (e.g., teens, seniors, wheelchair users), and robot (e.g., NAO, Moxi). It would also be interesting to gather information about participant demographics (e.g., age, ethnicity, citizenship, gender), and experience or attitude towards robotic technologies. Furthermore, conducting studies in a closed space to give participants a bit more comfort in sharing personal data, and informing them about their data security in our secure storage might produce more positive or interesting results. Another consideration in future studies is to have the robot respond based on participants' shared data as a two-way interaction, since valuable responses tying to their mental health can be impactful~\cite{facebook}.  
%
%
\printbibliography

@inproceedings{facebook,
author = {Bazarova, Natalya N. and Choi, Yoon Hyung and Schwanda Sosik, Victoria and Cosley, Dan and Whitlock, Janis},
title = {Social Sharing of Emotions on Facebook: Channel Differences, Satisfaction, and Replies},
year = {2015},
publisher = {Association for Computing Machinery},
address = {New York, NY, USA},
numpages = {11},
location = {Vancouver, BC, Canada},
series = {CSCW '15}
}

@article{SARTORIUS2007810,
title = {Stigma and mental health},
journal = {The Lancet},
volume = {370},
number = {9590},
pages = {810-811},
year = {2007},
author = {Norman Sartorius}
}

@article{brown2017discussing,
  title={Discussing out-of-pocket expenses during clinical appointments: an observational study of patient-psychiatrist interactions},
  author={Brown, Gregory D and Hunter, Wynn G and Hesson, Ashley and Davis, J Kelly and Kirby, Christine and Barnett, Jamison A and Byelmac, Dmytro and Ubel, Peter A},
  journal={Psychiatric Services},
  volume={68},
  number={6},
  pages={610--617},
  year={2017},
  publisher={Am Psychiatric Assoc}
}

@article{pella2020pediatric,
  title={Pediatric anxiety disorders: a cost of illness analysis},
  author={Pella, Jeffrey E and Slade, Eric P and Pikulski, Paige J and Ginsburg, Golda S},
  journal={Journal of abnormal child psychology},
  volume={48},
  number={4},
  pages={551--559},
  year={2020},
  publisher={Springer}
}

@article{aboujaoude2020digital,
  title={Digital interventions in mental health: current status and future directions},
  author={Aboujaoude, Elias and Gega, Lina and Parish, Michelle B and Hilty, Donald M},
  journal={Frontiers in psychiatry},
  volume={11},
  pages={111},
  year={2020},
  publisher={Frontiers}
}

@article{taylor2020digital,
  title={Digital technology can revolutionize mental health services delivery: The COVID-19 crisis as a catalyst for change},
  author={Taylor, C Barr and Fitzsimmons-Craft, Ellen E and Graham, Andrea K},
  journal={International Journal of Eating Disorders},
  volume={53},
  number={7},
  pages={1155--1157},
  year={2020},
  publisher={Wiley Online Library}
}

@article{digthe,
author = {Howells, Annika and Eiroa Orosa, Francisco Jose and Ivtzan, Itai},
year = {2015},
month = {01},
pages = {},
title = {Putting the 'app' in Happiness: A Randomised Controlled Trial of a Smartphone-Based Mindfulness Intervention to Enhance Wellbeing},
volume = {17},
journal = {Journal of Happiness Studies}
}

@inproceedings{Wo_niak_2021,
	
	year = 2021,
	month = {may},
  
	publisher = {{ACM}
},
  
	author = {Pawe{\l} W. Wo{\'{z}}niak and Jakob Karolus and Florian Lang and Caroline Eckerth and Johannes Schöning and Yvonne Rogers and Jasmin Niess},
  
	title = {Creepy Technology:What Is It and How Do You Measure It?},
  
	booktitle = {Proceedings of the 2021 {CHI} Conference on Human Factors in Computing Systems}
}

@article{McWhorter2020CreepyTA,
  title={Creepy Technologies and the Privacy Issues of Invasive Technologies},
  author={Rochell R. McWhorter and Elisabeth E. Bennett},
  journal={Research Anthology on Privatizing and Securing Data},
  year={2020}
}

@article{moodwall,
author = {Liu, Xuan and Pan, Mingtian and Li, Jia},
year = {2018},
month = {11},
pages = {},
title = {Does Sharing Your Emotion Make You Feel Better? An Empirical Investigation on the Association Between Sharing Emotions on a Virtual Mood Wall and the Relief of Patients' Negative Emotions},
volume = {25},
journal = {Telemedicine and e-Health},
doi = {10.1089/tmj.2017.0327}
}

@Article{robot,
AUTHOR = {Onyeulo, Eva Blessing and Gandhi, Vaibhav},
TITLE = {What Makes a Social Robot Good at Interacting with Humans?},
JOURNAL = {Information},
VOLUME = {11},
YEAR = {2020},
NUMBER = {1},
ARTICLE-NUMBER = {43}
}

@article{bishop,
  title={Social robots: The influence of human and robot characteristics on acceptance},
  author={Bishop, Laura and van Maris, Anouk and Dogramadzi, Sanja and Zook, Nancy},
  journal={Paladyn, Journal of Behavioral Robotics},
  volume={10},
  number={1},
  pages={346--358},
  year={2019},
  publisher={Sciendo}
}

@article{boucenna2014interactive,
  title={Interactive technologies for autistic children: A review},
  author={Boucenna, Sofiane and Narzisi, Antonio and Tilmont, Elodie and Muratori, Filippo and Pioggia, Giovanni and Cohen, David and Chetouani, Mohamed},
  journal={Cognitive Computation},
  volume={6},
  number={4},
  pages={722--740},
  year={2014},
  publisher={Springer}
}

@article{elin,
author = {Björling, Elin and Rose, Emma and Davidson, Andrew and Ren, Rachel and Wong, Dorothy},
year = {2020},
month = {01},
pages = {},
title = {Can We Keep Him Forever? Teens’ Engagement and Desire for Emotional Connection with a Social Robot},
volume = {12},
journal = {International Journal of Social Robotics}
}

@article{sharestress,
author = {Ling, Honson and Björling, Elin},
year = {2020},
month = {02},
pages = {133-158},
title = {Sharing Stress With a Robot: What Would a Robot Say?},
volume = {1},
journal = {Human-Machine Communication},
doi = {10.30658/hmc.1.8}
}

@inbook{inbook,
author = {Joshi, Swapna and Collins, Sawyer and Kamino, Waki and Gomez, Randy and Sabanovic, S.},
year = {2020},
month = {11},
pages = {440-452},
title = {Social Robots for Socio-Physical Distancing}
}

@ARTICLE{humano,
  
AUTHOR={Abubshait, Abdulaziz and Wiese, Eva},   
	 
TITLE={You Look Human, But Act Like a Machine: Agent Appearance and Behavior Modulate Different Aspects of Human–Robot Interaction},      
	
JOURNAL={Frontiers in Psychology},      
	
VOLUME={8},           
	
YEAR={2017}
}

@inproceedings{lbr,
author = {Karim, Raida and Zhang, Yufei and Alves-Oliveira, Patr\'{\i}cia and Bj\"{o}rling, Elin A. and Cakmak, Maya},
title = {Community-Based Data Visualization for Mental Well-Being with a Social Robot},
year = {2022},
publisher = {IEEE Press},
booktitle = {Proceedings of the 2022 ACM/IEEE International Conference on Human-Robot Interaction},
pages = {839–843},
numpages = {5},
location = {Sapporo, Hokkaido, Japan},
series = {HRI '22}
}

@article{schomerus2006public,
  title={Public beliefs about the causes of mental disorders revisited},
  author={Schomerus, Georg and Matschinger, Herbert and Angermeyer, Matthias C},
  journal={Psychiatry research},
  volume={144},
  number={2-3},
  pages={233--236},
  year={2006},
  publisher={Elsevier}
}

@misc{wiggers_2017, 
title={Knightscope's security robots: What they are, and how they work}, url={https://www.digitaltrends.com/cool-tech/knightscope-robots-interview/}, 
journal={Digital Trends}, 
publisher={Digital Trends}, 
author={Wiggers, Kyle}, 
year={2017}, 
month={Apr}}

@article{humanoid,
title = {Development of a New Mobile Humanoid Robot for Assisting Elderly People},
journal = {Procedia Engineering},
volume = {41},
pages = {345-351},
year = {2012},
note = {International Symposium on Robotics and Intelligent Sensors 2012 (IRIS 2012)},
author = {Zulkifli Mohamed and Genci Capi}
}

@Article{robotics8030054,
AUTHOR = {Nocentini, Olivia and Fiorini, Laura and Acerbi, Giorgia and Sorrentino, Alessandra and Mancioppi, Gianmaria and Cavallo, Filippo},
TITLE = {A Survey of Behavioral Models for Social Robots},
JOURNAL = {Robotics},
VOLUME = {8},
YEAR = {2019},
NUMBER = {3},
ARTICLE-NUMBER = {54}
}
\end{document}